\documentclass[10pt,conference,a4paper]{IEEEtran}
\IEEEoverridecommandlockouts
\usepackage{amsmath}
\usepackage{svg}
\usepackage{mathtools}
\usepackage{subfigure}
\usepackage{multirow}
\usepackage{arydshln}
\usepackage{graphicx}
\usepackage{epstopdf}
\usepackage{color}
\usepackage{amssymb}
\usepackage{balance}
\usepackage{url}
\usepackage{amssymb}
\usepackage{multirow}
\usepackage{arydshln}
\usepackage{threeparttable}
\usepackage{color}
\usepackage{algorithm}
\usepackage{algorithmic}
\usepackage{bm}
\usepackage{mathrsfs}
\usepackage{caption}
\usepackage{mathtools}
\usepackage{textcomp}
\usepackage{xcolor}
\usepackage{fancyhdr}
\usepackage{booktabs}
\usepackage{graphicx}
\usepackage{authblk}
\usepackage{multirow}
\usepackage{hyperref}

\def\BibTeX{{\rm B\kern-.05em{\sc i\kern-.025em b}\kern-.08em
    T\kern-.1667em\lower.7ex\hbox{E}\kern-.125emX}}

\begin{document}
\renewcommand\arraystretch{1.5}
\thispagestyle{empty}
\pagestyle{empty}

\title{A Unifying Tensor View for Lightweight CNNs}
\author{Jason Chun Lok Li*}
\author{Rui Lin*\thanks{*JL, and RL contributed equally to this work. }}
\author{Jiajun Zhou}
\author{Edmund Yin Mun Lam}
\author{Ngai Wong}
\affil{Department of Electrical and Electronic Engineering, The University of Hong Kong, Hong Kong}

\maketitle

\begin{abstract}
Despite the decomposition of convolutional kernels for lightweight CNNs being well studied, existing works that rely on tensor network diagrams or hyperdimensional abstraction lack geometry intuition. This work devises a new perspective by linking a 3D-reshaped kernel tensor to its various slice-wise and rank-1 decompositions, permitting a straightforward connection between various tensor approximations and efficient CNN modules. Specifically, it is discovered that a pointwise-depthwise-pointwise (PDP) configuration constitutes a viable construct for lightweight CNNs. Moreover, a novel link to the latest ShiftNet is established, inspiring a first-ever shift layer pruning strategy which allows nearly $50\%$ compression with $<1\%$ drop in accuracy for ShiftResNet.
\end{abstract}

\begin{IEEEkeywords}
Convolutional Neural Network, Tensor Decomposition, Shift Layer, Pruning
\end{IEEEkeywords}

\section{Introduction}

Lightweight deep neural networks (DNNs) are essential for edge artificial intelligence (AI), where DNNs operate on resource-limited hardware. This necessitates careful consideration of design constraints such as computation, storage, throughput and power consumption. Consequently, numerous efficient convolutional neural network (CNN) architectures have been developed, including prominent examples like MobileNet~\cite{MobileNetV2}, EfficientNet~\cite{EfficientNet}, and ShiftNet~\cite{Wu2018ShiftAZ}. Depthwise separable (DS) convolution~\cite{MobileNetV2, EfficientNet, Xception2017}, which replaces regular convolutional (CONV) kernels, has emerged as an effective solution, as it significantly reduces storage and computation requirements with little or no loss in output accuracy. This paper systematically and analytically demonstrates that these benefits arise from the underlying decomposition and approximation schemes of the reshaped CNN kernel tensor.

Matrix or tensor decomposition is a popular and powerful approach for compressing CNNs~\cite{cpdICLR2015,Tucker2,Astrid2018RankSO}. Ref.~\cite{cpdICLR2015} is among the first to propose utilizing the canonical polyadic decomposition (CPD) for compressing a regular CONV layer whose nature is a 4D kernel tensor. Specifically, the 4D kernel tensor is broken down into two depthwise (DW) layers and two pointwise (PW) layers using CPD. Ref.~\cite{Astrid2018RankSO} also employs CPD but treats the CONV layer as a reshaped 3D tensor, leading to a further reduction in the number of parameters. However, it only tests this method with the older-generation AlexNet with few CONV layers, using an ad hoc and non-scalable progressive layer-wise decomposition and fine-tuning scheme. Tucker decomposition is used in~\cite{Tucker2}, which decomposes a CONV layer into two PW layers and a regular CONV layer with smaller input and output channels. Notably, all these works are related to efficient CNNs~\cite{MobileNetV2, EfficientNet} containing DS or bottleneck layers. Ref.~\cite{einconv} enumerates various CNN decompositions using tensor network diagrams. However, the high-dimensional and abstract notations obscure the geometry and visual intuition associated with different DS variants. Although previous works have employed tensor network diagrams to represent higher-dimensional tensors and their decompositions into low-rank factors~\cite{cpdICLR2015, Tucker2, su2018tensorial}, the geometric view is lost, and the underlying arithmetic intuition becomes abstract at 4D and beyond. To address this issue, we demonstrate that it is possible to retain a 3D view and provide highly intuitive interpretations for various tensorized convolutions.

On another front of compacting CNNs, shift operations~\cite{Wu2018ShiftAZ} offer an ideal solution for edge devices due to their zero parameters and FLOPs. We accommodate shift layers in the CPD framework as \emph{one-hot} kernels, leading to a novel channel pruning scheme for shift layers. In fact, while tensors and various efficient CNN implementations are not new, it is \emph{the first time} they are brought under a highly visualizable, unifying tensor umbrella to provide an intuitive illustration of their origin of success. The arithmetic under various decompositions further inspires effective ways to compress the model. The main contributions of this work include \textbf{(i)} A reshaped kernel tensor view that unifies different approximation schemes and naturally spins off various efficient CNN architectures; \textbf{(ii)} The first to link CPD to hardware-efficient shift layers, leading to a \emph{first-of-its-kind} channel pruning scheme for shift layers.

\begin{figure*} [th]
    \centering
    \includegraphics[width=\textwidth]{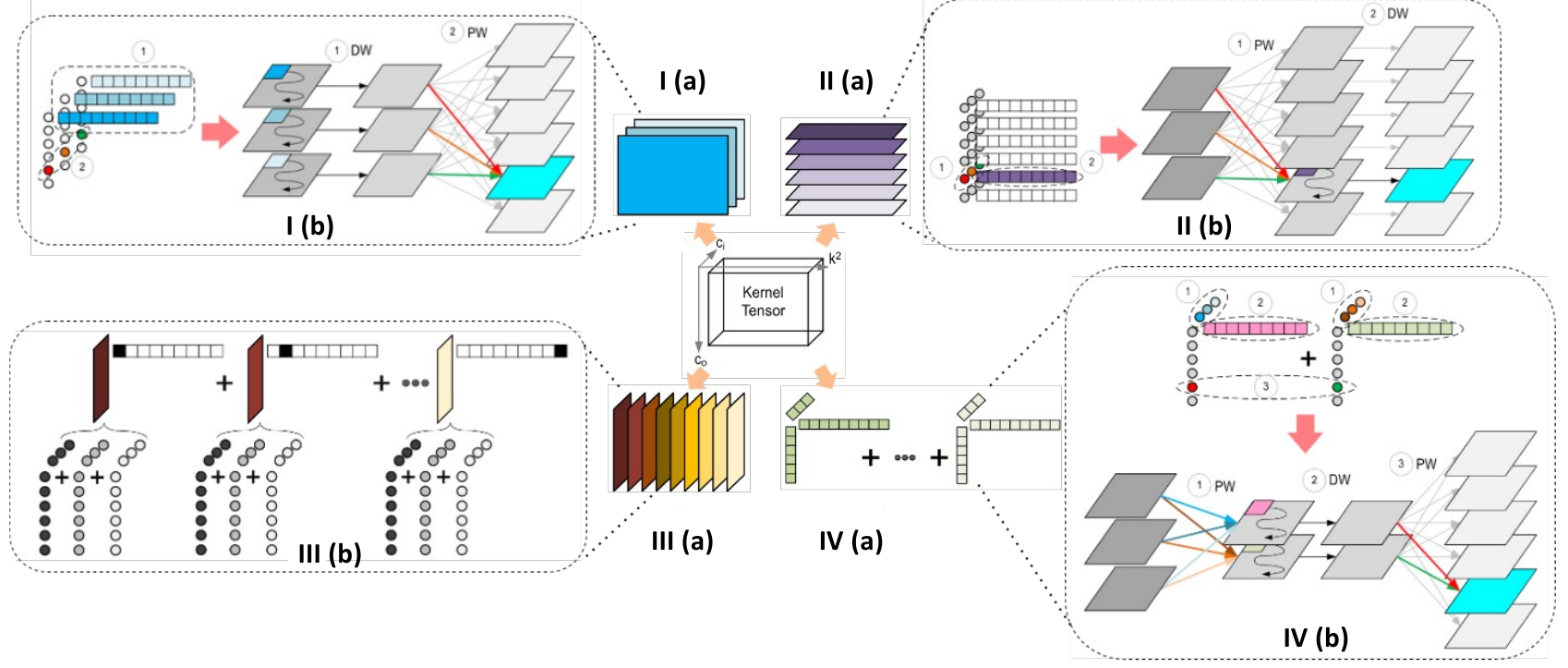}
    \vspace{-4mm}
    \caption{The reshaped kernel tensor and its different views: \textbf{I(a)} frontal slices; \textbf{II(a)}  horizontal slices; \textbf{III(a)} lateral slices; \textbf{IV(a)} CPD. We slightly abuse notations and use $c_o$, $c_i$ and $k^2$ for both axis and index labels. Assume $\left[|c_i|,|c_o|,|k|^2\right]=[3,6,9]$. \textbf{I(b)} Approximating frontal slices with rank-1 terms translates into the standard DS convolution (viz. DW + PW, abbr. DP). \textbf{II(b)} Approximating horizontal slices with rank-1 terms translates into an ``inverted'' depthwise separable convolution (viz. PW + DW, abbr. PD). \textbf{III(b)} Lateral slices positioned with one-hot kernel filters translate into a shift layer (viz. PW + Shift + PW) convolution. \textbf{IV(b)} Approximating the kernel tensor with multiple rank-1 CPD terms (CPD rank $r_{cp}=2$ here) translates into a linear bottleneck (viz. PW + DW + PW, abbr. PDP).}
    \label{fig:four_in_one}
\end{figure*}

\section{Kernel Tensor and its Approximation}
\label{sec:tensor_view}
To begin with, we instantiate the original 4-way kernel tensor of size  $\left[|c_i|,|c_o|,|k|,|k|\right]$ and reshape it into a 3-way format of size $\left[|c_i|,|c_o|,|k^2|\right]$ (cf. the center of Fig.~\ref{fig:four_in_one}), where $c_i, c_o$ denote the indices of input and output channels, and $k$ represents the width/height of the kernel window, while $|\circ|$ denotes the dimension/cardinality. Next, we enumerate various decompositions from different viewpoints on the 3-way kernel, which in turn generates different efficient CNN architectures. To our knowledge, such unifying visual interpretation in Fig.~\ref{fig:four_in_one} is presented only for the first time in the literature.

\subsection{Frontal Slices}
We assume $\left[|c_i|,|c_o|,|k^2|\right]=[3,6,9]$ for illustrative purpose. In Fig.~\ref{fig:four_in_one}.I(a), each slice or matrix can be decomposed into a sum of multiple rank-1 terms with decreasing dominance, e.g., by singular value decomposition (SVD). Fig.~\ref{fig:four_in_one}.I(b) depicts a rank-1 approximation to each frontal slice. Here we draw the spatial axis vectors of length $|k^2|$ with rectangular bars to differentiate them from other axes (using balls) to highlight their window sliding nature along the input channels they act on. In Fig.~\ref{fig:four_in_one}.I(b), we fix a particular output channel, e.g., $|c_o|:=6$, for illustration. It can be seen that there is one kernel filter along each $c_i$ index, corresponding to a DW convolution along each input channel. Once the DW convolution is performed, the 3 convolved output slices are produced which are then contracted by the 3 colored balls along the $c_i$ axis, namely, a PW operation to obtain the $|c_o|:=6$ channel. Such sequence of operations cannot be scrambled as the DW kernels are tied with each $c_i$ index, and need to be applied on each input channel first. The number of CNN weight parameters involved is $(|c_o|+|k^2|)|c_i|$ which equals 45 in this example. We abbreviate this CNN replacement by DP (viz. DW+PW) which coincides with the standard DS scheme.

\subsection{Horizontal Slices}
Similarly, we study Fig.~\ref{fig:four_in_one}.II(a) with its corresponding rank-1 slices shown in  Fig.~\ref{fig:four_in_one}.II(b). As before, we set $|c_o|:=6$, then it can be seen that there are 6 kernels along each $c_o$ index, each corresponding to a DW convolution operating on the PW ($1\times 1$)-contracted slices along the $c_i$ axis by the colored balls. Again, such sequence of operations cannot be swapped, since the 6 DW kernels are tied with each $c_o$ index, and need to be applied for producing each output channel. The number of weight parameters in this setting is $(|c_i|+|k^2|)|c_o|$ which equals 72 in our example. We call this PW+DW operation the PD scheme in this work, which can be seen as an ``inverted'' DS convolution. 

\subsection{Canonical Polyadic Decomposition}
We detour to the CPD view in Fig.~\ref{fig:four_in_one}.IV(a) before returning to the lateral slices, which makes a more natural order of illustration. Referring to  Fig.~\ref{fig:four_in_one}.IV(b) and borrowing from previous sections, we can segregate the contraction/convolution into three stages, namely: 1) PW along $c_i$ to generate a number of slices equal to the CPD rank $r_{cp}$ (i.e., number of CPD terms), 2) DW along these slices, and 3) PW along the desired $c_o$. This view corresponds to the celebrated bottleneck layer~\cite{MobileNetV2}, whose residual variant is depicted in the upper part of Fig.~\ref{fig:shift_layer}. Whether it is a standard or inverted bottleneck is determined by $r_{cp}$, which decides the center DW block being wider or thinner than its ``entrance'' and ``exit'' PW layers. We name this PW+DW+PW combination \textit{PDP}. In this example, the number of weight parameters is $(|c_i|+|c_o|+|k^2|)r_{cp}$ which equals 36. Apparently, this number depends heavily on $r_{cp}$ which we can tune for different tradeoffs between complexity and representation capacity.

\subsection{Lateral Slices}

Finally, we turn to Fig.~\ref{fig:four_in_one}.III(a) for the lateral slices. It is clear we can view the original tensor as a slice-wise aggregation by positioning them at the appropriate $k^2$-axis index via a one-hot kernel vector (cf.  Fig.~\ref{fig:four_in_one}.III(b)). Likewise, each slice can be decomposed into multiple rank-1 terms, all sharing the same one-hot kernel vector. For instance, the leftmost term in Fig.~\ref{fig:four_in_one}.III(b) can spin off into three CPD terms, resembling those in  Fig.~\ref{fig:four_in_one}.IV(b). The one-hot nature of the kernel vector effectively performs simple shifting, as advocated in~\cite{Wu2018ShiftAZ}. As such, we draw equivalence of lateral slices to their ShiftNet counterparts. If we use $r_{cp}$ to denote the total number of 3-way rank-1 terms selected, say, according to their importance characterized by their singular values, then the number of weights equals $(|c_i|+|c_o|)r_{cp}$.

\section{Shift Layer Pruning}
\label{sec:shift_prune}
\begin{figure}[t]
\centering
  \includegraphics[width=.45\textwidth]{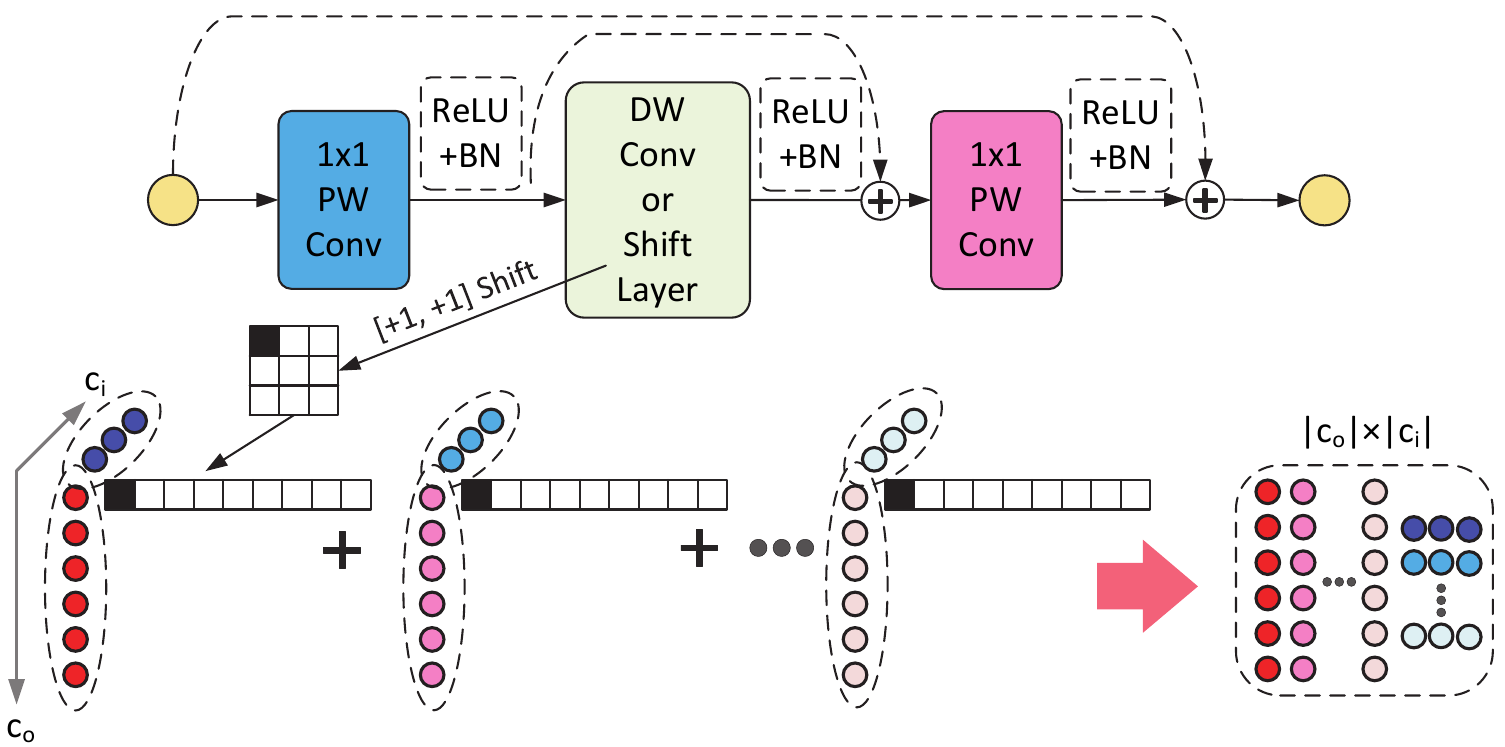}
  \caption{(Upper) A generic PW+DW/Shift+PW block where the dashed-line shortcuts and nonlinearity blocks are optional. (Lower) Assuming a shift layer, one can collect $c_o$-$c_i$ mode rank-1 terms of the same shift and sum them for the derivation of principal components.}
  \label{fig:shift_layer}
\end{figure}

Based on the novel view of Section~\ref{sec:tensor_view}, if a pretrained ShiftNet is available, then one way to do pruning of its shifted channels is to collect rank-1 terms corresponding to the same shift, and then add up the $c_o$-$c_i$ mode rank-1 terms into a matrix (cf. lower part of Fig.~\ref{fig:shift_layer}). Then SVD is performed on this $|c_o|\times |c_i|$ matrix to decide the number of dominant terms for retention and the number of terms to drop. To the best of our knowledge, this is the first-ever shift layer pruning scheme.

\begin{figure}[t]
    \centering
    \includegraphics[scale=0.3]{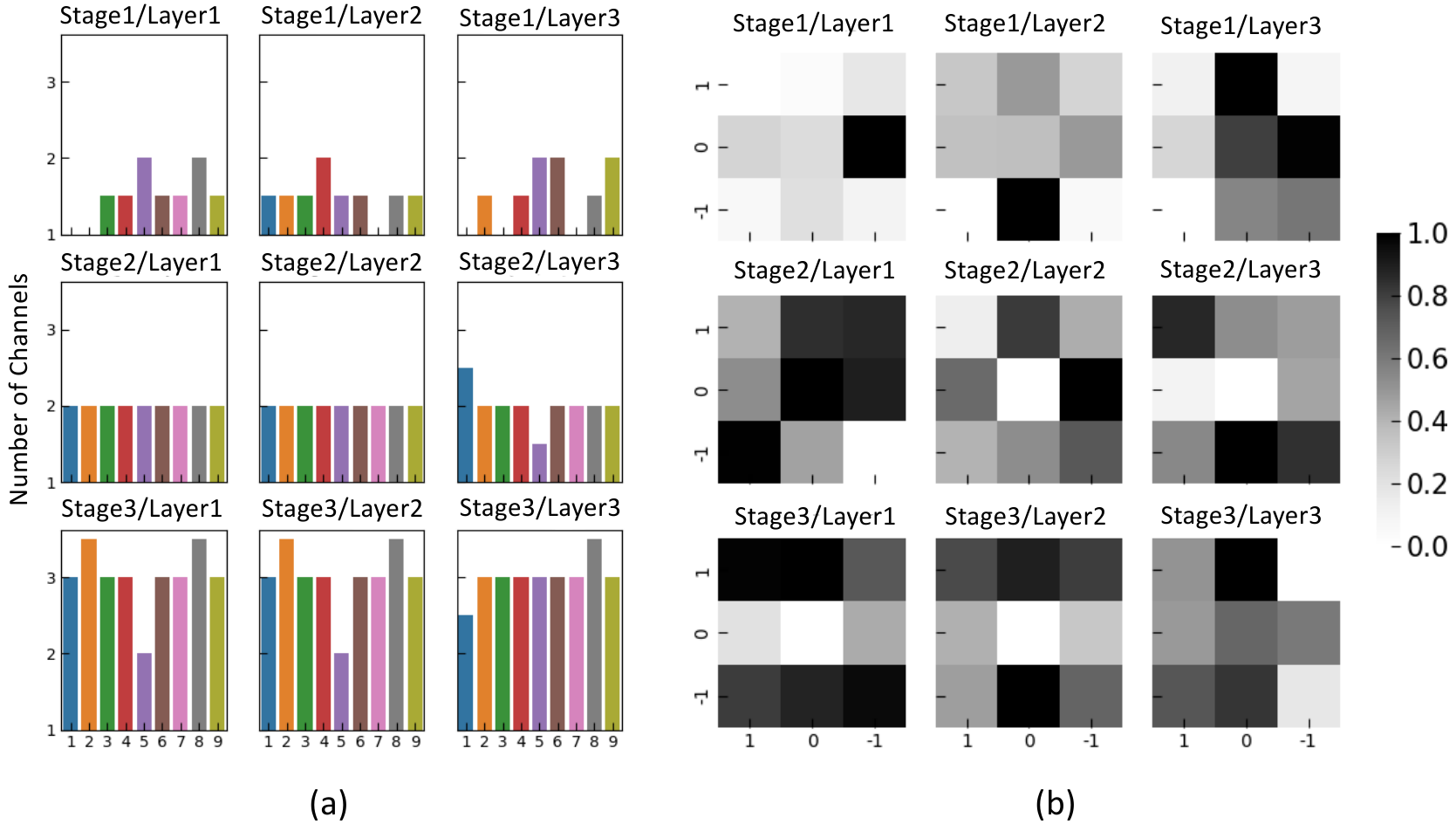}
    \vspace{-1mm}
    \caption{Shift distribution of ShiftResNet-20 trained on CIFAR-10 after uneven shift pruning with $\phi=1/16$. There are three stages in ShiftResNet-20, each having three shift layers. (a) The number of channels in each shift group; (b) The relative importance of each shift in various layers based on their singular values (which are summed at each position and normalized within each layer).} \label{fig:shift_distribution}
\end{figure}

Specifically, under the same shift, the $c_o$-$c_i$ rank-1 terms are summed up into a $|c_o|\times |c_i|$ matrix for principal component analysis, e.g., via SVD. Then, the most dominant terms are retained as the new PW filters on the $c_i$ and $c_o$ modes. As an illustrative example, suppose for two channels with the same shift, the $c_i$-mode $1\times 1$ filters are $v, \hat{v}\in \mathbb{R}^{|c_i|}$ whose corresponding $c_o$-mode $1\times 1$ filters are $u, \hat{u}\in \mathbb{R}^{|c_o|}$, respectively. Then the summed matrix is $\left[ {\begin{array}{*{20}c}
   u & \hat{u}  \\
\end{array}} \right]\left[ {\begin{array}{*{20}c}
   {v^T }  \\
   {\hat{v}^T }  \\
\end{array}} \right]$. Suppose the flattened input channel matrix is $X$ and there exists nonlinearity (say, ReLU) after the first PW layer. Then, if $\hat{v}\approx v$, whether in terms of cosine or Euclidean distance, the mapped channel matrices ${\rm ReLU}(v^TX)\approx {\rm ReLU}(\hat{v}^TX)$, especially when the feature dimension is high. The same argument also holds when $\hat{u}\approx u$  which, together with $\hat{v}\approx v$, implies the above matrix would be close to rank-1 and its principal $c_i$ and $c_o$ filters can be obtained via SVD and further fine-tuned through backpropagation. Subsequently, when the entrance and exit PW layers have ``close'' PW filters, then they can be consolidated into fewer principal filters. Note that our pruning approach is different from ~\cite{Chen2019AllYN}, where shift operations are sparsified through a loss regularizer, while holding the size of the intermediate feature map unchanged. In contrast, we treat the one-hot shift kernel vector as an on/off switch carrying information of its associated $c_i$-$c_o$ rank-1 term. Once pruned, the entire 3-way rank-1 term is dropped, so the feature map size and representation capacity both reduce.

Here, we define a new hyperparameter dubbed pruning ratio $\phi=\epsilon_{new}/\epsilon_{old}$, where the notion of expansion ratio $\epsilon$ is borrowed from~\cite{Wu2018ShiftAZ} which specifies the ratio of bottleneck's channel size to the output channel size of the exit PW. Obviously, $\phi$ controls the degree of compression, allowing for a tradeoff between accuracy and model size. There are two strategies to select the principal filters. The first involves comparing singular values within the same shift and preserving the same amount of dominant terms in each group. Concretely, suppose we have a $3\times 3$ DW shift layer, then there are 9 shift groups each corresponding to one direction of shift (including zero shift). This method guarantees an equal number of channels in each shift group. Another approach is to compare singular values across all shift groups and retain the most dominant terms. Fig.~\ref{fig:shift_distribution} shows the shift distribution under ``uneven'' shift pruning, where missing bars in Fig.~\ref{fig:shift_distribution}(a) illustrate that the whole shift group is pruned. 

\section{Experiments}
\label{sec:experiments}

\subsection{VGG-16}
\label{exp:vgg}

Here approximations with different tensor views discussed in Section~\ref{sec:tensor_view} are realized by replacing \emph{every} CONV layer of a VGG-16. A baseline VGG-16 is first trained on CIFAR-10 for 200 epochs. The pretrained CONV kernels are then decomposed into various configurations followed by 300 epochs of fine-tuning. Table~\ref{tab:vgg_pw_dw} shows the results of using DP and PD schemes to decompose the kernel tensors. The PD configuration has a slightly better performance as compared to DP with the same degree of compression. Here the compression is measured by the model-wise compression rate (CR) which will be used throughout. Surprisingly, random initialization leads to a higher test accuracy in both cases as compared to SVD initialization. We believe that the reason is approximation using a single rank-1 term for each tensor slice is insufficient. Table~\ref{tab:vgg_cpd}, on the other hand, highlights the significance of CPD initialization when every CONV layer in the VGG-16 is turned into a PDP configuration comprising $r_{cp}$ DW filters. VGG-16 with CPD-initialized kernels outperforms its random-initialized counterpart by a significant margin, ranging from $11\%$ to $14\%$. By gradually raising the CPD rank ($r_{cp}$) from $4$ to $32$, classification accuracy rises from $62.48\%$ to $89.73\%$ with only a slight decrease in compression ratio ($99.65\%$ vs $98.12\%$). Indeed, such PDP slimming of a dense network has a distinct advantage over DP and PD, which achieves an additional $\approx 6\times$ parametric reduction than the latter two. Utilizing CPD initialization for PDP training, the final accuracy is mostly restored.

\begin{table}[t]
\setlength{\tabcolsep}{5.5mm}{
\centering
\scriptsize
\caption{VGG-16 with SVD vs randomly initialized DP/PD kernels on CIFAR-10.}
\label{tab:vgg_pw_dw}
\begin{tabular}{cccc}
\toprule
\multirow{2}{*}{Method} & \multirow{2}{*}{CR (\%)} & \multirow{2}{*}{\#Params(M)} & Acc (\%) \\
\cline{4-4}
~ & ~ & ~ & Random/SVD\\
\midrule
DP & $88.53$&$1.69$ & $\mathbf{92.81}/92.70$  \\
\midrule
PD & $88.49$&$1.69$ & $\mathbf{93.72}/93.54$  \\
\bottomrule
\end{tabular}}
\centering
\scriptsize
\caption{VGG-16 with CPD vs randomly initialized PDP kernels on CIFAR-10.}
\label{tab:vgg_cpd}
\setlength{\tabcolsep}{5.3mm}{
\begin{tabular}{cccc}
\toprule
\multirow{2}{*}{Rank $r_{cp}$} & \multirow{2}{*}{CR (\%)} & \multirow{2}{*}{\#Params(M)} & Acc (\%) \\
\cline{4-4}
~ & ~ & ~ & Random/CPD\\
\midrule
4 & $99.65$&$0.05$ & $51.31/\mathbf{62.48}$  \\
\midrule
8 & $99.44$&$0.08$ & $61.63/\mathbf{72.27}$  \\
\midrule
16 & $99.00$&$0.15$ & $65.06/\mathbf{86.09}$  \\
\midrule
32 & $98.12$&$0.28$ & $75.19/\mathbf{89.73}$  \\
\bottomrule
\end{tabular}}
\end{table}
\begin{table}[b]
\centering
\scriptsize
\caption{Shift Layer Pruning. Baseline accuracies are $93.37\%$@CIFAR-10, $71.47\%$@CIFAR-100.}
\label{tab:shift_cifar}
\setlength{\tabcolsep}{1.8mm}{
\begin{tabular}{cccccc}
\toprule
\multirow{2}{*}{$\phi$} & \multicolumn{2}{c}{CIFAR-10} & ~ & \multicolumn{2}{c}{CIFAR-100} \\
\cline{2-3}
\cline{5-6}
\noalign{\smallskip}
\noalign{\smallskip}
~   & CR (\%)  & Acc (Even/Uneven) (\%)  & ~ & CR (\%)  & Acc (Even/Ueven) (\%) \\ 
\midrule
$\frac{1}{2}$ & $49.18$& $92.81/\mathbf{92.88}$ & ~ & $48.17$ & $70.48/\mathbf{70.58}$ \\
\midrule
$\frac{1}{4}$ & $73.78$& $91.41/\mathbf{91.50}$ & ~ & $72.25$ & $67.16/\mathbf{67.62}$ \\
\midrule
$\frac{1}{8}$ & $86.07$& $89.09/\mathbf{89.11}$ & ~ & $84.30$ & $62.21/\mathbf{63.13}$ \\
\midrule
$\frac{1}{16}$ & $92.22$& $85.57/\mathbf{85.84}$ & ~ &$90.32$ & $\mathbf{57.25}/56.83$ \\
\bottomrule
\end{tabular}}
\end{table}

\subsection{Shift Layer Pruning}
\label{exp:shift_pruning}
To validate the idea of shift layer pruning as described in Section~\ref{sec:shift_prune}, experiments are conducted on CIFAR datasets. ShiftResNets~\cite{Wu2018ShiftAZ}  with an expansion ratio $\epsilon=9$ are employed. A shift module in the ShiftResNet consists of a $3\times3$ DW shift layer surrounded by two PW layers, interleaved with BN and nonlinearity.  By varying $\epsilon_{new}$, we obtain a set of pruning ratios $\phi\in\{\frac{1}{2},\frac{1}{4},\frac{1}{8},\frac{1}{16}\}$. ShiftResNet-20 is used as the backbone. Baseline models are first obtained by training on target datasets for 200 epochs, followed by another 200 epochs of fine-tuning. As seen from Table~\ref{tab:shift_cifar}, shift layer pruning achieves nearly $50\%$ compression with $< 1\%$ accuracy drop for both CIFAR-10 ($92.88\%$ vs $93.37\%$) and CIFAR-100 ($70.58\%$ vs $71.47\%$).

\section{Conclusion}
In this paper, a novel 3D tensor exposition of various CNN kernel approximations is presented (cf. Fig.~\ref{fig:four_in_one}), revealing their close bonds to separable pointwise (PW) and depthwise (DW) convolutions. This perspective further enables the development of a first-ever channel pruning scheme for the zero-flop and hardware-efficient shift layers. We believe such a unifying tensor view can establish itself as a fundamental framework for CNN approximation theory, and serve as an essential basis for discovering new and efficient CNN architectures.
\section{Acknowledgment}
\label{sec:acknowlege}
This work was supported in part by the General Research Fund (GRF) project 17209721, and in part by the Theme-based Research Scheme (TRS) project T45-701/22-R of the Research Grants Council (RGC), and partially by ACCESS – AI Chip Center for Emerging Smart Systems, sponsored by InnoHK funding, Hong Kong SAR.

\tiny
\bibliographystyle{ieeetr}
\bibliography{bare_conf}

\begin{thebibliography}{10}

\bibitem{MobileNetV2}
M.~Sandler, A.~G. Howard, M.~Zhu, A.~Zhmoginov, and L.-C. Chen, ``Mobilenetv2:
  Inverted residuals and linear bottlenecks,'' pp.~4510--4520, 2018.

\bibitem{EfficientNet}
M.~Tan and Q.~Le, ``{E}fficient{N}et: Rethinking model scaling for
  convolutional neural networks,'' in {\em International Conference on Machine
  Learning}, vol.~97 of {\em Proceedings of Machine Learning Research},
  pp.~6105--6114, PMLR, 09--15 Jun 2019.

\bibitem{Wu2018ShiftAZ}
B.~Wu, A.~Wan, X.~Yue, P.~H. Jin, S.~Zhao, N.~Golmant, A.~Gholaminejad, J.~E.
  Gonzalez, and K.~Keutzer, ``Shift: A zero flop, zero parameter alternative to
  spatial convolutions,'' pp.~9127--9135, 2018.

\bibitem{Xception2017}
F.~Chollet, ``Xception: Deep learning with depthwise separable convolutions,''
  in {\em IEEE Conference on Computer Vision and Pattern Recognition (CVPR)},
  July 2017.

\bibitem{cpdICLR2015}
V.~Lebedev, Y.~Ganin, M.~Rakhuba, I.~V. Oseledets, and V.~S. Lempitsky,
  ``Speeding-up convolutional neural networks using fine-tuned
  cp-decomposition,'' in {\em International Conference on Learning
  Representations}, 2015.

\bibitem{Tucker2}
Y.~Kim, E.~Park, S.~Yoo, T.~Choi, L.~Yang, and D.~Shin, ``Compression of deep
  convolutional neural networks for fast and low power mobile applications,''
  in {\em International Conference on Learning Representations}, 2016.

\bibitem{Astrid2018RankSO}
M.~Astrid, S.-I. Lee, and B.-S. Seo, ``Rank selection of {CP}-decomposed
  convolutional layers with variational {B}ayesian matrix factorization,'' {\em
  International Conference on Advanced Communication Technology (ICACT)},
  pp.~347--350, 2018.

\bibitem{einconv}
K.~Hayashi, T.~Yamaguchi, Y.~Sugawara, and S.~ichi Maeda, ``Einconv: Exploring
  unexplored tensor network decompositions for convolutional neural networks,''
  in {\em Advances in Neural Information Processing Systems}, vol.~32,
  pp.~5552--5562, Curran Associates, Inc., 2019.

\bibitem{su2018tensorial}
J.~Su, J.~Li, B.~Bhattacharjee, and F.~Huang, ``Tensorial neural networks:
  Generalization of neural networks and application to model compression,''
  {\em arXiv preprint arXiv:1805.10352}, 2018.

\bibitem{Chen2019AllYN}
W.~Chen, D.~Xie, Y.~Zhang, and S.~Pu, ``All you need is a few shifts: Designing
  efficient convolutional neural networks for image classification,''
  pp.~7234--7243, 2019.

\end{thebibliography}

\end{document}